\documentclass[a4paper]{article}
\usepackage{INTERSPEECH2020}
\usepackage{graphicx}
\usepackage{amsmath}
\usepackage{multirow}
\usepackage{float}

\title{MOTS: Multiple Object Tracking for General Categories Based On Few-Shot Method}
\name{Xixi Xu, Chao Lu$^*$, Liang Zhu, Xiangyang Xue, Guanxian Chen, Qi Guo, Yining Lin, Zhijian Zhao}
\address{
	Fudan University, Shanghai, China; Supremind Co., Ltd., Shanghai, China
}
\email{Xixi Xu: 18210240227@fudan.edu.cn, Chao Lu: luchao256258@126.com}

\begin{document}

\maketitle

\begin{abstract}
Most modern Multi-Object Tracking (MOT) systems typically apply REID-based paradigm to hold a balance between computational efficiency and performance. In the past few years, numerous attempts have been made to perfect the systems. Although they presented favorable performance, they ware constrained to track specified category. Drawing on the ideas of few shot method, we pioneered a new multi-target tracking system, named MOTS, which is based on metrics but not limited to track specific category. It contains two stages in series: In the first stage, we design the self-Adaptive-matching module to perform simple targets matching, which can complete 88.76\% assignments without sacrificing performance on MOT16 training set. In the second stage, a Fine-match Network was carefully designed for unmatched targets. With a newly built TRACK-REID data-set, the Fine-match Network can perform matching of 31 category targets, even generalizes to unseen categories’.

\end{abstract}

\noindent\textbf{Index Terms}: Multiple Object Tracking, General Categories, New Dataset

\section{Introduction}

The association of targets that encounter occlusion is still a key challenge in Multi-Object Tracking systems. For example, IOU Tracker \cite{bochinski2018extending} mainly depends on spatial overlap to track, which was susceptible to a relatively large spatial gap between the last position and the newly appear position of the obscured target. Although Deep SORT \cite{wojke2017simple} can handle the association of obscured targets by utilizing cosine distance, they only target at a specified category, tracking objects of a new category requires a new carefully trained model. Therefore, it leads to a novel research direction, how to equip the metric-based Multi-Object Tracking systems with the ability to track targets beyond specified categories, even generalize to unseen categories. Actually, we find that in single object tracking, it is common sense for trackers to track any arbitrary objects given the annotation in the first frame, even its category never appear in training. After further analysis, we discover that the essences of single object tracking is to learn discriminative yet robust features for comparison in a data driven way. Thus, to enhance the generalization of Multi-Object Tracking system, what really matters is to find a proper way for comparison. To our best knowledge, this is the first attempt to address multiple object tracking for general categories by designing a novel yet robust similarity metric that originate from few-show learning. Utilizing our newly built TRACK-REID dataset, it turns out to be a promising way for generalizing.

Usually, in Multi-Object Tracking system, most targets are at a distance from each other, named easy targets, which can be easily associated by using spatial overlap. On the contrary, if the targets that missing detection for a long time or encounter occlusion, it was challenging for association as their locations usually change a lot when reappeared, named hard targets, which requires metric comparison. Thus, we design a self-Adaptive-matching module based on IOU matrix to distinguish the easy targets and perform association, which outperforms the Hungarian algorithm based on IOU matrix.

MOTS is an online frame by frame Multi-Object Tracking system, which not all the temporarily disappeared tar-gets require association. For the targets that normally move out the sight, we introduce an mv-aware trajectories deletion module to perform quick deletion, so as to reduce computation.

To sum up, the contributions of our work are three folds.

(1) We introduce a metric learning method that originate from few-shot learning to Multi-Object Tracking systems and newly built the TRACK-REID data-set. Our MOTS can thus track multi objects of arbitrary categories. 

(2) We proposed a two stage Multi-Object Tracking system. The first stage utilize self-Adaptive-matching module based on IOU matrix to perform easy targets’ association and the second stage apply metric learning to associate hard targets. 

(3) We further propose an mv-aware trajectories deletion module to perform quick deletion on normally disappear targets, which is also a plug-and-play module.

\section{Related Works}
%\begin{figure*}[!htbp]
%	\centering
%	\includegraphics[height=6cm]{fig_Architecture}
%	\caption{Architecture of the Fine-Match Network}
%	\label{fig:example}
%\end{figure*}

Multi-Object Tracking methods always follow the tracking-by-detection paradigm. They can be categorized into offline tracking and online tracking according to their processing mode. Offline tracking utilizes a batch of frames to process associations, the representative works includes \cite{zhang2008global, maksai2017non, wen2019learning}. Online tracking also named sequential tracking, the association is handled in a step-wise manner. Our work belongs to online tracking. We briefly introduce three main types online tracking methods below.

The first type is Multi-Object Tracking systems that em-ploy single object trackers. Xiang et al.\cite{xiang2015learning} formulated the online Multi-Object Tracking problem as a MDP task and learn a similarity function in a reinforcement learning fashion. They employ a single object tracker as an agent in MDP to track the target, which can benefit from the strength of online single object tracking methods. Chu et al.\cite{chu2017online} designed a CNN-based single object tracker to perform Multi-Object Tracking. They further apply spa-tial-temporal attention mechanism to handle the drift problem. Zhu et al.\cite{zhu2018online} combined on-the-shell single object tracker ECO-HC\cite{danelljan2017eco} and data association methods for online Multi-Object Tracking. They introduced a cost-sensitive tracking loss and propose Dual Matching Attention Networks to further enhance the performance. Feng et al.\cite{feng2019multi} presented a unified Multi-Object Tracking system where a single object tracking sub-net extracted short term cues and a REID sub-net got long term cues. Bochinski et al.\cite{bochinski2018extending} utilized single object tracker to address missing detection to assist their IOU tracker.

The second type is Multi-Object Tracking systems that combine motion models and appearance models. Bewley et al.\cite{bewley2016simple} explored a simple pragmatic approach for online and real-time Multi-Object Tracking. By adapting Kalman Filter and Hungarian algorithm, SORT updates at a rate of 260Hz, outperform other trackers by a large margin. After that, they further integrated deep appearance features to enhance the performance and proposed Deep SORT\cite{wojke2017simple}. Although the cosine distance can measure appearance similarity in some degree, the fixed metric that compare features element-wise lacks the flexibility and can’t perform well in all categories. On the contrary, we employ a deep non-linear metric that can be trained in a data driven way and generalize to unseen categories\cite{sung2018learning}. Besides, Joint Detection and Embedding (JDE) was proposed by Wang et al.\cite{wang2019towards} which became the first (near) real-time Multi-Object Tracking system instead of real-time association methods only. This was realized by incorporate the appear-ance-embedding model into a single-shot detector.

Other kind of online Multi-Object Tracking systems including IOU-based methods and beyond. Bochinski et al.\cite{bochinski2017high} proposed a simple yet efficient IOU tracker based on the assumption that none or only few “gaps” in the detection. Thus, they associate the targets simply by spatial overlap and gain a high tracking speed.

\section{Method}

As mentioned above, the MOTS consists of two stages. Easy targets’ association can be tackled by a plug-and-play self-Adaptive-matching module based on IOU matrix in the first stage. Then, the unmatched targets enter into the second stage and a trainable similarity metric was adopted to perform association.

\subsection{Self-Adaptive-matching module based on IOU matrix }

The first stage is expected to categorize the targets into easy targets as well as hard ones no matter in what conditions (such as fixed cameras or unregularly moving cameras) and further perform associations on easy targets. In the spirit of simple and efficient, the module mainly consists of four steps.
Step 1, provided N detection,$dec_{0},…,dec_{N-1}$,and M existing targets with their new bounding boxes,$track_{0},…,track_{M-1}$, predicted by Kalman Filter, on the current frame. The IOU matrix $\text {iou}_{j, i} \in[0,1]^{M \times N}$ is then calculated as the intersection-over-union (IOU) distance between every detection and target pair.

\begin{figure*}[!htbp]
	\centering
	\includegraphics[height=6cm]{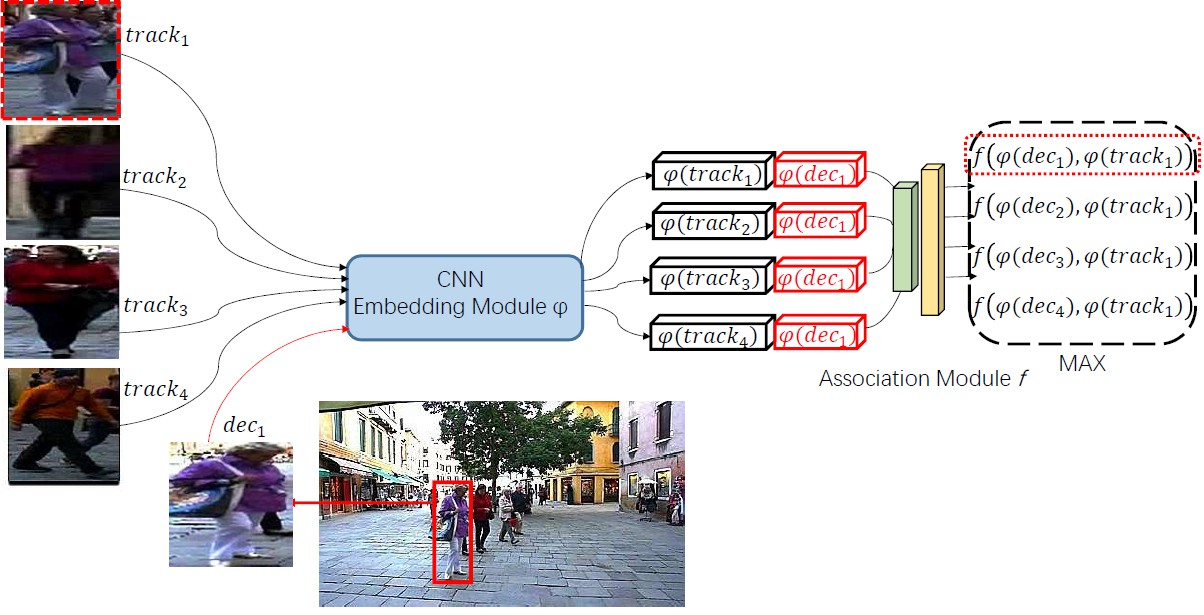}
	\caption{Architecture of the Fine-Match Network}
	\label{fig:example}
\end{figure*}

\begin{equation}
\label{eq1}
i o u_{j, i}=\frac{\text {Area}\left(\text {track}_{j}\right) \cap \text {Area}\left(\text {dec}_{i}\right)}{\text {Area}\left(\text {track}_{j}\right) \cup \text {Area}\left(\text {dec}_{i}\right)}
\end{equation}

Where $\text {Area}\left(\text {track}_{j}\right)$ are the area of ${track}_{j}$, and $\text {Area}\left(\text {dec}_{i}\right)$ represents the area of ${dec}_{i}$.

Step 2, in order to adapt to targets’ various motion states at different times, we design a basic IOU value ${base}_{j,t}$ for each targets $j$ at time $t$.
\begin{equation}
\label{eq2}
base_{j, t}=\frac{\sum_{t-t_{n_{1}}}^{t} i o u_{j, t}}{t_{n 1}}
\end{equation}

Where $ t $ represents the current moment. We take the latest ${t}_{n1}$ moments into consideration. ${t}_{n1}$ usually be 5. ${iou}_{j,t}$ is the IOU distance between the ${track}_{j}$ and its matched detection at $ t $ moment. The matching procession is performed through step3 and step 4, ${iou}_{j,t}$ is calculated following formula (5).

The basic IOU matrix follows two principles. First, each target has its own basic IOU value ${base}_{j,t}$. Second, the value should responds quickly to changes in the tracking target's motion states at different moments. Therefore, the values ${base}_{j,t}$ indicates the most likely IOU distance between adjacent frames for the corresponding target.

We then update the base IOU value ${base}_{j,t}$ of each targets at current moment and determinate the IOU threshold for matching according to the basic IOU value ${base}_{j,t}$. Differ from IOU Tracker that adopts fixed IOU threshold for all targets, our IOU threshold is adaptive for each targets and updates timely. This idea originate from two aspects. First, different targets have various motion states; they may at different locations of a moving camera, leading to a big discrepancy on IOU distances among matched pairs. Besides, even for a same target, the matched IOU distances may differ a lot at different moments due to the changes in motion velocity or the unregularly moving camera.

Step 3, the clamp and normalization of the IOU matrix. If the ${iou}_{j,i}$ is smaller than the minimal threshold ${throd}_{min}$, it is believed the probability that the pairs belongs to a same target would be extremely small, as the IOU distance of two unrelated targets in a crowded area would be close to that value. So we zero them out to eliminate their interference.
\begin{equation}
\label{eq3}
i o u_{j, i}=\left\{\begin{array}{cc}
i o u_{j, i} & , \text {if } i o u_{j, i} \geq t h r o d_{\min } \\
0 & , e l s e
\end{array}\right.
\end{equation}

As various targets’ motion states differs on a same frame, we apply normalization to regulate them into a uniform scale. To be concrete, each IOU value ${iou}_{j,i}$ in the matrix have to be divided by the corresponding basic IOU value ${base}_{j,t}$, and get $ {iou\_norm}_{j, i} \in[0,2.5] $. It presents the matching degree between ${dec}_{i}$ and ${track}_{j}$, taking the latest ${t}_{n}$ moments into account. We find that most matched ${iou\_norm}_{j,i}$ are around 1, smaller than 1 indicates the matching become worse, otherwise means that become better.
\begin{equation}
\label{eq4}
i o u_{\text {norm}_{j, i}}=\left\{\begin{array}{cc}
\frac{i o u_{j, i}}{b a s e_{j, t}} & , \text {if } i o u_{j, i} \geq t \text {hrod}_{\min } \\
0 & , \text {else}
\end{array}\right.
\end{equation}

Step 4, the matching strategy based on normalized IOU matrix. Instead of directly apply Hungarian algorithm as other methods, we design a simple yet efficient module that is more suitable to multi object tracking, which outperforms Hungarian algorithm based on IOU matrix. First, we traverse the normalized IOU matrix, if ${iou\_norm}_{j,i}$ is the maximum in both row and column and $ \text {iou\_norm}_{j, i} \geq m a t c h_{\min } $, the ${track}_{j}$ and ${dec}_{i}$ become a candidate matching pairs, where ${match}_{min}$ is the minimum threshold and we set it to 0.85. Then, to exclude the cases that one detection matching multiple targets or one target matching multiple detection, we further traverse the $ j^{th} $ row and $ i^{th} $ column to get the secondary maximum ${iou\_norm}_{j,ix}$ and ${iou\_norm}_{jy,i}$. If they both smaller than ${match}_{min}$, we ultimately match ${track}_{j}$ to ${dec}_{i}$.The pairs that violate any of the above conditions will be regarded as hard pairs and enter the second stage.

For matched pairs ${track}_{j}$ and ${dec}_{i}$, their IOU distance ${iou}_{j,t}$ at $ t $ moment is computed as follows.
\begin{equation}
\label{eq5}
i o u_{j, t}=\left\{\begin{array}{cc}
0 & , \text { if } \mathrm{t}=t_{0} \\
i o u_{j i} & , \text { else } i f \ i o u_{j i} \geq t h r o d_{m i n} \\
i o u_{j, t-1} & , e l s e
\end{array}\right.
\end{equation}

Where ${t}_{0}$ is the initialization moment.

%\begin{figure*}[!htbp]
%	\centering
%	\includegraphics[height=5cm]{3}
%	\caption{Architecture of the Fine-Match Network}
%	\label{fig:example}
%\end{figure*}

\subsection{Fine-match Network Architecture}
After the association in the first stage, there are much fewer targets enter the second stage, which make it possible to employ a sophisticated yet powerful distance measure within acceptable computation. The basic idea of this stage originate from few-shot learning that to train a deep similarity measure which can generalize to unseen categories. Therefore, it can break the limitation and track multiple objects beyond specified categories.

It formally contains three data-sets: a training set, a support set and a testing set. The label space for the support set and the testing set is the same, which is disjoint with the training set. For a C-way K-shot task, its support set consists of C classes with K samples from each. A trained classifier is able to appoint the test samples to corresponding category in support set. Here in MOT system, we regard the unmatched targets,$track_{0},\dots,track_{M_{2}-1}$,as ${M}_{2}$ categories in the support set and construct a ${M}_{2}$Way 1Shot task. Regard the unmatched detection,$dec_{0},\dots,dec_{N_{2}-1}$,as ${N}_{2}$ samples in the test set.

Fine-match Network consists of an embedding module $ \varphi $ and an association module $ f $. The former one is excepted to extract deep features from both the test sample and each categories in the support set, generating $ \varphi\left(\operatorname{dec} _{i}\right)$ and $ \varphi\left(\operatorname{tra c k} _{0}\right), \ldots, \varphi\left(\operatorname{tra c k}_{M_{2}-1}\right) $, respectively. Then, each $ \varphi\left(\operatorname{dec} _{i}\right)$  is concated to all $ \varphi\left(\operatorname{track} _{j}\right)$  and enter the association module, which is responsible for similarity measure. For each detection $ {dec} _{i} $, the $ {track} _{j} $ that get the maximum $ f\left(\varphi\left(d e c_{i}\right), \varphi\left(\text { track }_{j}\right)\right) $ is believed to be the most suitable one for association. The architecture of the whole fine-match network is as follows.

We adopt the MSE loss function for Fine-match Network training. The similarity for matched pairs are1, others are 0.

\begin{equation}
\label{eq6}
\begin{split}
\text { Loss }=&\sum_{i=0}^{N_{2}-1} \sum_{j=0}^{M_{2}-1}\\
&\left(f\left(\varphi\left(\operatorname{dec}_{i}\right), \varphi\left(\operatorname{track}_{j}\right)\right)-1\left(y_{\operatorname{dec}_{i}}==y_{\text {track}_j}\right)\right)^{2}
\end{split}
\end{equation}

%\begin{figure*}
%	\centering
%	\includegraphics[height=5cm]{3}
%	\caption{Architecture of the Fine-Match Network}
%	\label{fig:example}
%\end{figure*}

\begin{table}[!htbp]
	\begin{center}
		\caption{TRACK-REID dataset composition. S-dataset refers to sub-datasets. VID-Val refers to the val set of VID. ‘proc’ refers to way of procession. ‘R’ refers to resize. ‘C’ refers to crop. ‘C-n’ refers to Categories number. ‘I-n’ refers to iden-tification number. ‘P-n’ refers to picture numbers.} 
		\label{table1}
		\begin{tabular}{lllll}
			\hline\noalign{\smallskip}
			s-dataset & proc & C-n & I-n & P-n\\
			\noalign{\smallskip}
			\hline
			\noalign{\smallskip}
			VeRI & R&	1	&576	&37778\\
			VRIC&	R&	1&	2138&	51459\\
			Market1501&	R&	1&	554&	11631\\
			MSMT17&	R&	1&	3890&	110749\\
			PKU&	R&	1&	114&	1824\\
			Duke&	R&	1&	695&	16463\\
			VID-Val&	C\&R&	30&	621	&273505\\
			OTB100&	C\&R&	$ \approx $2&	100&	58943\\
			\hline
		\end{tabular}
	\end{center}
\end{table}

\begin{table*}[!htbp]
	\begin{center}
		\caption{Above methods all adopt a same detector. ’SA’ refers to self-adaptive-matching module.’H’ refers to Hungarian.‘M-track’ refers to the ratio of total matched tracks. ‘M-det’ refers to the ratio of total matched detection. The ‘base’ indicates that only metric association is applied for multi object tracking. That is, for MOTS, only employ the fine-match network for association. We alternately apply self-adaptive-matching strategy and Hungarian algorithm to Deep SORT and MOTS. Total-matchs-num refers to the number of targets that processed in the first stage. Total-detects-num refers to the total number of detection from all frames. The value of match-detect-ratio equals to total-matchs-num/total-detects-num. Total-tracks-num refers to the total numbers of tracking targets from all frames. The value of match-track-ratio equals to total-matchs-num/total-tracks-num.} 
		\label{table2}
		\begin{tabular}{lllllllll}
			\hline\noalign{\smallskip}
			method&	structure&MOTA$\uparrow$&MT$\uparrow$&ML$\downarrow$&IDS$\downarrow$&M-track (\%)$\uparrow$&	M-det (\%)$\uparrow$\\
			\noalign{\smallskip}
			\hline 
			\noalign{\smallskip}
			\multirow{3}{*}{DeepSORT} &base&60.1&166&84&733&$\backslash$&	$\backslash$\\
			&base\&H&59.8&170&84&1239&64.30&89.89\\
			&base\&SA&\textbf{60.1}&166&86&1207&65.66&88.79\\
			\hline
			\multirow{3}{*}{MOTS} &base&57.1&163&85&2428&$\backslash$&$\backslash$\\
			&base\&H&58.9&155&91&1254&58.55&89.89\\
			&base\&SA&\textbf{59.1}&162&87&1227&43.03&87.83\\
			\hline
		\end{tabular}
	\end{center}
\end{table*}

\subsection{TRACK-REID dataset}
Although we adopt a trainable similarity measure from few-shot learning, we can’t employ its dataset directly to train our module. The normal dataset in few-shot learning like mini-ImageNet is for classification instead of identification. The essential difference in tracking and classification lies in that tracking have to match detections to the targets of the same identification, not just of a same category. While for identification of other categories except person and vehicle, there still lack corresponding datasets. Therefore, we newly build a TRACK-REID dataset to tackle the challenge.

The TRACK-REID consists of three sub-set. One is the Val set from VID \cite{russakovsky2015imagenet}, which contains 30 categories. We find that for each short videos in VID data-set, the target is exactly an identification. We further crop the target from each frames according to the groundtruth bounding boxes. Consider that the miniImageNet is big enough to train a few-shot network, we only combine the val set instead of the whole set of VID to our TRACK-REID. And the later experiments also con-firm that TRACK-REID is enough to train a robust network and hold a balance between the efficiency and performance. In addition, as the OTB100\cite{wu2015object} benchmark in single object tracking contains more real tracking conditions, we also add them into our TRACK-REID with the same procession. The last sub-set is collected from several person re-identification and vehicle re-identification data-sets, including MSMT17\cite{wei2018person}, PKU\cite{ma2016orientation}, Duke\cite{zheng2017unlabeled, gou2017dukemtmc4reid}, Market1501\cite{zheng2015scalable} and VRIC\cite{kanaci2018vehicle}, VeRI\cite{liu2016large}, respectively. The newly built dataset is as follows.

\subsection{mv-aware trajectories deletion method}
The design of tracking object management module follows the general operation, which contains three types: initialize, update and delete. The initialize and update follow the general operation. If the target appears continuously for twice, a new trajectory will be initialized, and the new lo-cation will be used to update the trajectory in each frame. In the delete operation, differs from the usual way of set-ting a fixed threshold for all targets to judge whether they disappear, we design a method to distinguish the targets needed deletion according to the movement displacement. The lost targets can be categorized into two types. One are the temporarily lost targets that lack detection due to the occlusion, blurring and other factors. The others are the normal lost targets that moving out of the view. These two types need to be treated differently. The former will usually reappear and should be equipped with a longer life span, while the latter tends not to reappear and should be deleted in time to save computation resources.

Therefore, an mv-aware trajectories deletion method is proposed to quickly distinguish two kinds of vanishing targets and set corresponding deletion threshold. We record and update the average moving velocity of the latest 5 frame, e.g.,$ t_{n 2}=5 $, for each tracking target. Its velocity is the distance between the center of its current position and that of its previous position. When the distance between the target and any image boundary is larger than 2 times of the moving velocity in that direction, the target is determined to be the former type, and it is deleted only when no new matching target appears in the continuous $ throd_{del1}=30 $ frames. Otherwise, the target is determined to be the latter type, and will be deleted if no matching target appears in the continuous $ throd_{del2}=3 $ frames. Experiments show that this simple discriminative-deletion module can significantly reduce the number of targets that are sent to Fine-Match Network and thus reduce time consuming.

\section{Experiments}
\begin{table*} [!htbp]
	\begin{center}
		\caption{Experiments on generalization.}
		\label{table3} 
		\begin{tabular}{llllll}
			\hline\noalign{\smallskip}
			Training set&Bus(\%)&Lizard(\%)&Lion(\%)&Tiger(\%)&Motorcycle(\%)\\
			\noalign{\smallskip}
			\hline
			\noalign{\smallskip}
			TRACK-REID-&81.09&90.92&82.12&83.55&81.22\\
			TRACK-REID&83.30&91.83&83.81&85.07&83.21\\
			\hline
		\end{tabular}
	\end{center}
\end{table*}

\begin{table} 
	\begin{center}
		\caption{Experiments on better training set.‘VID-Val’ refers to the val set of VID. ‘VID-All’ refers to the whole set of VID. ‘VID-Val-REID’ refers to the combination of the val set of VID and REID data-sets mentioned above.}
		\label{table4} 
		\begin{tabular}{ll}
			\hline\noalign{\smallskip}
			Training set&Test set: vehicle (\%)\\
			\noalign{\smallskip}
			\hline
			\noalign{\smallskip}
			VID-Val&80.06\\
			VID-All&83.17\\
			VID-Val-REID&95.37\\
			Track-REID&95.22\\
			\hline
		\end{tabular}
	\end{center}
\end{table}
\begin{table*} [!htbp]
	\label{table5} 
	\begin{center}
		\caption{’B’ refers to base model without self-adaptive-matching module nor mv-aware-deletion module.’SA’ refers to self-adaptive-matching module. ’MA’ refers to mv-aware-deletion module. ‘M-det’ refers to the ratio of total matched detection. ‘M-track’ refers to the ratio of total matched tracks. This analyze the effect of adaptive-matching module and mv-aware-deletion module. ‘Base’ refers to MOTS without mv-aware-deletion module or self-adaptive-matching module.}
		
		\begin{tabular}{lllllllll}
			\hline\noalign{\smallskip}
			structure&	MOTA$\uparrow$&	IDF1$\uparrow$&	MT$\uparrow$&ML$\downarrow$&IDS$\downarrow$&M-det(\%)$\uparrow$&M-track(\%)$\uparrow$&FPS$\uparrow$\\
			\noalign{\smallskip}
			\hline
			\noalign{\smallskip}
			B&57.1&54.7&163&85&2428&$\backslash$&$\backslash$&7.5\\
			B\&MA&57.0&53.1&163&84&2555&0.00&3.72&7.8\\
			B\&SA&59.1&57.5&162&87&1227&87.83&43.03&23.2\\
			B\&SA\&MA&59.1&57.5&159&89&1126&87.84&62.08&26.4\\
			\hline
		\end{tabular}
	\end{center}
\end{table*}
\begin{table*} [!htbp]
	\label{table6}
	\begin{center}
		\caption{The comparison of three MOT systems. Above methods all adopt a same detector.}
		
		\begin{tabular}{lllllllll}
			\hline
			\noalign{\smallskip}
			method                                     & Detector & MOTA$\uparrow$ & IDF1$\uparrow$ & MT$\uparrow$ & ML$\downarrow$ & IDS$\downarrow$ & FPS$\uparrow$ & Catogories$\uparrow$ \\
			\noalign{\smallskip}
			\hline
			\noalign{\smallskip}
			DeepSORT & POI      & 61.4           & 62.2           & 32.8         & 18.2           & 781             & 17.4          & 1                    \\
			POI                                                                & POI      & 66.1           & 65.1           & 34.0         & 20.8           & 805             & 9.9           & 1                    \\
			MOTS(ours)                                                         & POI      & 60.2           & 61.1           & 30.7         & 20.2           & 1594            & 16.4          & \textgreater31       \\ \hline
		\end{tabular}
	\end{center}
\end{table*}

\subsection{Comparison of self-Adaptive-matching module and Hungarian algorithm base on IOU}
The proposed self-Adaptive-matching module based on IOU matrix embodies powerful robustness, which is insensitive to super-parameters. We experimentally set ${t}_{n}$ to 5, ${throd}_{min}$ to 0.4 and ${match}_{min}$  to 0.85 for all the experiments. Although thorough parameters’ tuning may gain better results, it beyond our discussion as this work focus on proposing a robust and with promising generalization method.

We corroborate the effectiveness of the self-adaptive-matching method by comparing it with Hun-garian algorithm. The experiments on MOT16 challenge training set are as Table \ref{table2}. From Table \ref{table2}, we can find that, in the deep SORT method, adopting the proposed self-adaptive-matching module can complete the association of 88.79\% detection and 65.66\% tracking targets, on the premise of maintaining a same MOTA. On the contrast, although the Hungarian algorithm can also match 89.89\% detection to 64.03\% tracking targets, it sacrifices 0.2\% of the MOTA. The 0.2\% reduction in MOTA usually occur when a detection have relatively high IOU distances with more than two targets. Moreover, in MOTS system, self-adaptive-matching method also outperform Hungarian algorithm by a 0.2\% margin with a relatively lower total-matchs-num. This validate that self-adaptive-matching method do have the ability to distinguish the easy targets and hard ones adaptively. 

\subsection{Experiments on generalization}
We adopt a simplified Resnet 256 as our embedding module, which is similar with\cite{mishra2017simple}. As for the association module, it contains 5 layers. The first layer consists of a convolution layer with 128 channels and the kernel size is 3*3, followed by a batch normalization layer and a relu layer. The second and the third layer’s structures are the same, containing a convolution layer with 64 channels and the kernel size is 3*3, followed by a batch normalization layer, a relu layer and a max pooling layer. The last two layers are linear layers.

For the training of the fine-match module, we apply the Adam optimizer with a learning rate of 0.0001. The episode based training are also utilized to mimic the few-shot learning setting. That is, in each training iteration, C categories with K samples from each were selected randomly from the training set to form a temporal support set. Another 5 samples were selected from these C categories to form a temporal test set. The temporal support set and test set form an episode and calculate the MSE loss. The training episodes range from 50000 to 70000 depending on the size of training data-set.

We randomly remove 5 categories from TRACK-REID to form a TRACK-REID- data-set for training, so as to corroborate the generalization of the fine-match network. The removed categories are then serve as the testing set. It shows that even the specified categories never appear in the training set, the fine-match network can still gain comparable results to the whole training set configuration.

To further explore a better way for enhancing the generalization of the fine-match network. We randomly divide each car data-sets VRIC and VeRI into a training set and a test set, with a ratio of 8:2. The test set is reserve for testing the network’s performance.

As showed above, a larger data-set benefit the performance. Besides, with more category relevant data can benefit a lot.

%\begin{table*} [!htbp]
%	\label{table5} 
%	\begin{center}
%		\caption{’B’ refers to base model without self-adaptive-matching module nor mv-aware-deletion module.’SA’ refers to self-adaptive-matching module. ’MA’ refers to mv-aware-deletion module. ‘M-det’ refers to the ratio of total matched detection. ‘M-track’ refers to the ratio of total matched tracks. This analyze the effect of adaptive-matching module and mv-aware-deletion module. ‘Base’ refers to MOTS without mv-aware-deletion module or self-adaptive-matching module.}
%		
%		\begin{tabular}{lllllllll}
%			\hline\noalign{\smallskip}
%			structure&	MOTA$\uparrow$&	IDF1$\uparrow$&	MT$\uparrow$&ML$\downarrow$&IDS$\downarrow$&M-det(\%)$\uparrow$&M-track(\%)$\uparrow$&FPS$\uparrow$\\
%			\noalign{\smallskip}
%			\hline
%			\noalign{\smallskip}
%			B&57.1&54.7&163&85&2428&$\backslash$&$\backslash$&7.5\\
%			B\&MA&57.0&53.1&163&84&2555&0.00&3.72&7.8\\
%			B\&SA&59.1&57.5&162&87&1227&87.83&43.03&23.2\\
%			B\&SA\&MA&59.1&57.5&159&89&1126&87.84&62.08&26.4\\
%			\hline
%		\end{tabular}
%	\end{center}
%\end{table*}

\subsection{Ablation study}
The parameters of part 3.3 can be set as follows: $ t_{n2}=5$, $ throd_{del1}=30$,$ throd_{del2}=3$. The effect on MOT16-train is as follows.

Fine-Match Network is able to measure more categories than REID. It can even generalize to categories unseen previously, while its measure ability is weaker than methods that specific to a single category. Self-adaptive matching strategy and mv-aware-deletion module can effectively reduce the number of targets sent into Fine-Match Network, which can not only reduce the computation, but also enhance the performance. From the Table 5, it can be seen that, with the application of self-adaptive matching module and mv-aware-deletion module, the MOTA in-creased from 57.1\% to 59.1\%, while the running speed also increased from 7.5fps to 26.4fps.

\subsection{Benchmark Evaluation}

The proposed MOTS system aims to track tracking multi objects that beyond specified categories, which is realized by Fine-Match Network. It has been verified in section 4.2. In addition to tracking targets up to 31 categories in TRACK-REID data-set, MOTS even have the ability to track unseen categories. It should be noted that, as metric measure based MOT system highly depend on detectors, MOTS does not have a great advantage on tracking multiple objects of normal categories, such as the pedestrians in MOT16 challenge. Unlike DeepSort and POI that are carefully design for the specific category, while our MOTS is a general tracker. The comparison results are on Table 6. A balance between performance on specified category and generalization is still worth exploring for the further work.
%\setlength{\tabcolsep}{4pt}
%\begin{table} 
%	\label{table6}
%\begin{center}
%\caption{The comparison of three MOT systems. Above methods all adopt a same detector.}
%
%\begin{tabular}{lllllllll}
%	\hline
%	\noalign{\smallskip}
%method                               & Detector & MOTA$\uparrow$ & IDF1$\uparrow$ & MT$\uparrow$ & ML$\downarrow$ & IDS$\downarrow$ & FPS$\uparrow$ & Catogories$\uparrow$ \\
%	\noalign{\smallskip}
%\hline
%\noalign{\smallskip}
%DeepSORT & POI      & 61.4           & 62.2           & 32.8         & 18.2           & 781             & 17.4          & 1                    \\
%	POI                                                       & POI      & 66.1           & 65.1           & 34.0         & 20.8           & 805             & 9.9           & 1                    \\
%	MOTS(ours)                                                & POI      & 60.2           & 61.1           & 30.7         & 20.2           & 1594            & 16.4          & \textgreater31       \\ \hline
%\end{tabular}
%\end{center}
%\end{table}
%\setlength{\tabcolsep}{1.4pt}

\section{Conclusions}
We dedicated to tackle the long-standing challenge in metric-based multi-object tracking, that is, to inherit a high performance of metric measure while break the limitation of tracking specified category. Our Fine-Match Network originate from few-shot learning can address the problem in some degrees. To our best knowledge, this is the first attempt to track multiple general objects with a Fine-Match Network and the newly build TRACK-REID data-set. The self-Adaptive-matching module that based on IOU matrix and the mv-aware trajectories deletion module are two plug-and-play module that can also be integrated in other MOT systems to further enhance the performance. We will continue to improve. The next step is to further improve the Fine-Match Network so that the MOTS system can approach performance of REID-based system infinitely when the number of hard targets is acceptable (usually less than 50) and still maintain the generalization ability. In fact, the self-adaptive matching method and mv-aware-deletion designed by us alleviate this problem implicitly by reducing the number of hard targets.

\bibliographystyle{IEEEtran}

\bibliography{mybib}

\end{document}